\documentclass{egpubl}
\usepackage{pg2024}

\SpecialIssueSubmission    % uncomment for submission to , special issue
\usepackage[T1]{fontenc}
\usepackage{dfadobe} 
\usepackage{amsmath}
\usepackage{amssymb}
\usepackage{multirow}

\usepackage{cite}  % comment out for biblatex with backend=biber
\BibtexOrBiblatex
\electronicVersion
\PrintedOrElectronic
\ifpdf \usepackage[pdftex]{graphicx} \pdfcompresslevel=9
\else \usepackage[dvips]{graphicx} \fi

\usepackage{egweblnk} 
\title[Semantic-guided Adversarial Diffusion Model for Self-supervised Shadow Removal]%
      {Semantic-guided Adversarial Diffusion Model for Self-supervised Shadow Removal}

\author[Ziqi Zeng, Chen Zhao, Weiling Cai \& Chenyu Dong]
{\parbox{\textwidth}{\centering Ziqi Zeng\thanks{These authors contributed equally to this work.}, Chen Zhao\footnotemark[1], Weiling Cai\thanks{Corresponding Author.}, Chenyu Dong
        }
        \\
{\parbox{\textwidth}{\centering School of Artificial Intelligence, Nanjing Normal University
       }
}
}

\begin{document}

% uncomment for using teaser
% \teaser{
%  \includegraphics[width=0.9\linewidth]{eg_new}
%  \centering
%   \caption{New EG Logo}
% \label{fig:teaser}
%}

\maketitle
%-------------------------------------------------------------------------
\begin{abstract}
   Existing unsupervised methods have addressed the challenges of inconsistent paired data and tedious acquisition of ground-truth labels in shadow removal tasks. However, GAN-based training often faces issues such as mode collapse and unstable optimization. Furthermore, due to the complex mapping between shadow and shadow-free domains, merely relying on adversarial learning is not enough to capture the underlying relationship between two domains, resulting in low quality of the generated images. To address these problems, we propose a semantic-guided adversarial diffusion framework for self-supervised shadow removal, which consists of two stages. At first stage a semantic-guided generative adversarial network (SG-GAN) is proposed to carry out a coarse result and  construct paired synthetic data through a cycle-consistent structure. Then the coarse result is refined with a diffusion-based restoration module (DBRM) to enhance the texture details and edge artifact at second stage. Meanwhile, we propose a multi-modal semantic prompter (MSP) that aids in extracting accurate semantic information from real images and text, guiding the shadow removal network to restore images better in SG-GAN. We conduct experiments on multiple public datasets, and the experimental results demonstrate the effectiveness of our method.
%-------------------------------------------------------------------------
\begin{CCSXML}
<ccs2012>
   <concept>
       <concept_id>10010147.10010371.10010382.10010383</concept_id>
       <concept_desc>Computing methodologies~Image processing</concept_desc>
       <concept_significance>500</concept_significance>
       </concept>
   <concept>
       <concept_id>10010147.10010178.10010224.10010240.10010241</concept_id>
       <concept_desc>Computing methodologies~Image representations</concept_desc>
       <concept_significance>500</concept_significance>
       </concept>
 </ccs2012>
\end{CCSXML}

\ccsdesc[500]{Computing methodologies~Image processing}
\ccsdesc[500]{Computing methodologies~Image representations}

\printccsdesc   
\end{abstract}  
%-------------------------------------------------------------------------
\section{Introduction}
In the realm of human perception, visual, auditory, and tactile senses collectively serve as conduits through which individuals interact with the world. Among these senses, vision stands out as the most intuitive and information-rich mode of perception. Visual perception occurs when light interacts with the retina, giving rise to visual experiences. Shadows, a natural consequence of obstructed light sources, play a crucial role in shaping the visual landscape. While shadows convey valuable cues regarding object shapes and light direction, their presence often complicates the semantic understanding of images in computer vision tasks such as image segmentation\cite{li2018learning} and object detection\cite{fang2019moving}. The shadow regions may be misclassified as objects or part of object thereof may significantly impair the accuracy and performance of these tasks. Consequently, shadow detection and removal are essential for enhancing the efficacy of computer-based visual tasks.
\begin{figure}[htb]
  \centering
  % the following command controls the width of the embedded PS file
  % (relative to the width of the current column)
  \includegraphics[width=.8\linewidth]{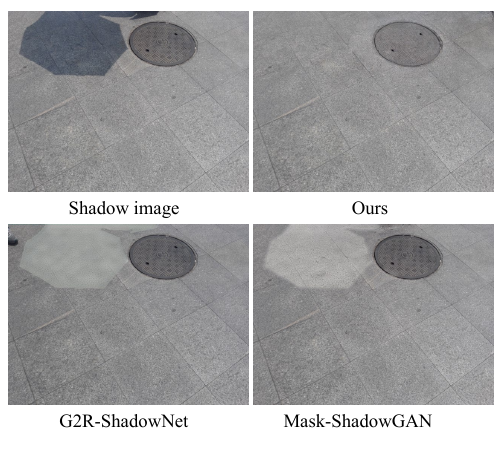}
  % replacing the above command with the one below will explicitly set
  % the bounding box of the PS figure to the rectangle (xl,yl),(xh,yh).
  % It will also prevent LaTeX from reading the PS file to determine
  % the bounding box (i.e., it will speed up the compilation process)
  % \includegraphics[width=.95\linewidth, bb=39 696 126 756]{sampleFig}
  %
  \caption{\label{fig:fig1-1}
           The shadow removal results of our method and other two GAN-based methods: G2R-ShadowNet\cite{liu2021shadow} and Mask-ShadowGAN\cite{2019Mask}. Gan-based methods have obvious shadow boundaries and artifacts.}
\end{figure} 

Early shadow removal methods\cite{2011Shadow} \cite{2006On} \cite{2013Paired} \cite{2013Fast}focus on exploring shadow removal in images according to different physical properties of shadows. Due to the insufficient accuracy and limitations of the underlying physical model, traditional physical model-based shadow removal algorithms are unable to effectively address shadows in complex real-world scenes. 

Learning-based methods\cite{2020CLA} \cite{cun2020towards}\cite{zhu2022efficient} typically train networks using paired shadow images and corresponding shadow-free images in fully supervised manner. However, inconsistencies exist in large-scale paired shadow removal datasets due to uncontrollable nature of outdoor illumination. Some methods have been proposed, which do not require paired data for training and generate supervisory signals through shadow generation using a cycle-consistent architecture. Nevertheless, the gap between synthetic and real images limits the full application of these methods in real-world shadow removal task. Although existing unsupervised methods have achieved notable results in shadow removal, GAN-based methods are susceptible to unstable optimization and mode collapse during training \cite{xiao2021tackling} \cite{zhao2024cycle}. Additionally, adversarial training alone is insufficient to fully learn the complex mapping between shadow and shadow-free domains. As shown in Figure~\ref{fig:fig1-1}, the resulting shadow-free images often exhibit visible boundary artifacts and lack detailed texture restoration, leaving room for improvement in visual quality.
\begin{figure}[htb]
  \centering
  % the following command controls the width of the embedded PS file
  % (relative to the width of the current column)
  \includegraphics[width=.8\linewidth]{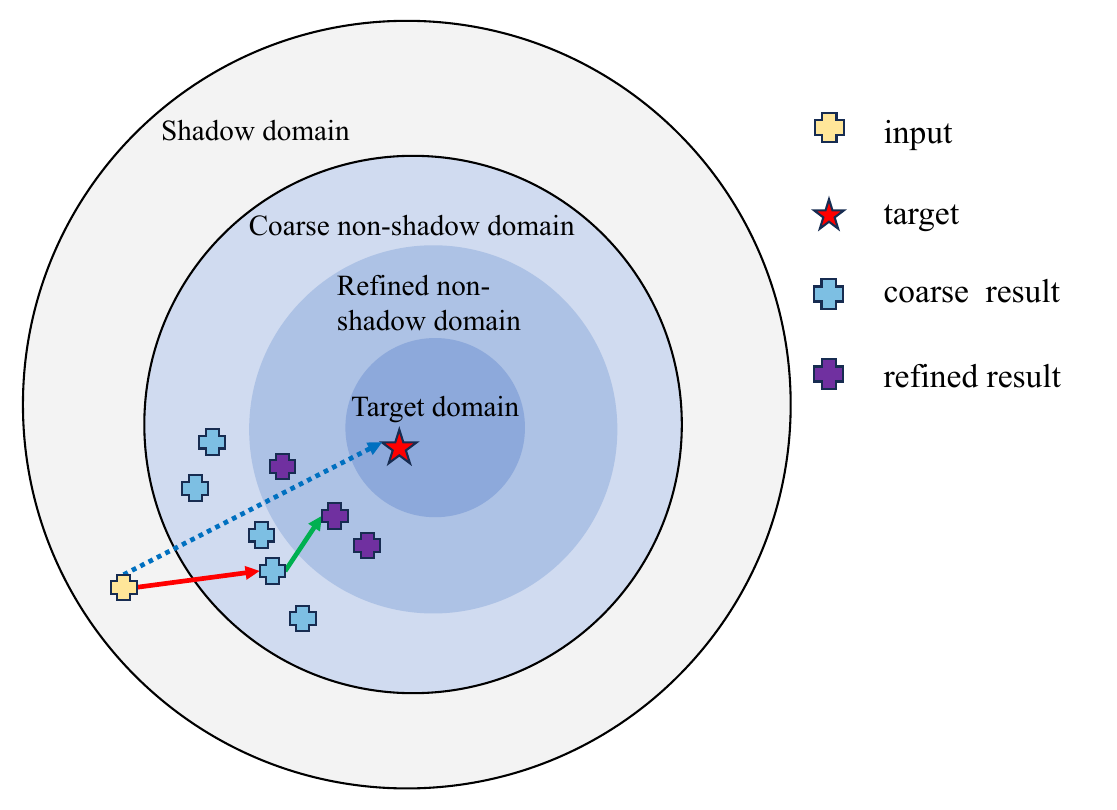}
  % replacing the above command with the one below will explicitly set
  % the bounding box of the PS figure to the rectangle (xl,yl),(xh,yh).
  % It will also prevent LaTeX from reading the PS file to determine
  % the bounding box (i.e., it will speed up the compilation process)
  % \includegraphics[width=.95\linewidth, bb=39 696 126 756]{sampleFig}
  %
  \caption{\label{fig:fig1}
           Visualization of different domains and domain transitions. The red arrow represents the previous adversarial generation process, and the green arrow represents the diffusion generation process.}
\end{figure} 

In recent years, diffusion models, e.g.the Denoising Diffusion Probabilistic Model (DDPM) \cite{ho2020denoising}, have achieved significant breakthroughs in visual generative tasks as a new branch of generative models \cite{zhou2023pyramid} \cite{li2023diffusion} \cite{zhou2024migc} \cite{lu2023tf} \cite{lu2024mace} \cite{rombach2022high} \cite{saharia2022palette}. Owing to their superior generative capabilities, many studies have explored the application of diffusion models in image restoration tasks to enhance texture recovery. ShadowDiffusion \cite{guo2023shadowdiffusion} proposes a diffusion sampling strategy to explicitly integrate the shadow degradation prior into the inherent iterative process of the dynamic mask sensing diffusion model. Liu \cite{liu2024recasting} reconstructs the local illumination of the shadow region by diffusion model. Although diffusion models provide a more stable training process and are more effective in capturing the pixel distribution of images in contrast to GANs. These diffusion-based methods still require paired data for training and are encumbered by the limitations inherent to the supervised paradigm. Consequently, we aim to explore self-supervised paradigm based on diffusion models.

As shown in Figure~\ref{fig:fig1}, we consider first using the GAN-based method to carry out coarse shadow removal, and in this process construct paired synthetic data through a cycle-consistent structure. Then we use diffusion model to process the paired data and refine the coarse results to make them closer to our target. Through this process, the diffusion model can be used in self-supervised shadow removal. In this paper, we propose a novel semantic-guided adversarial diffusion framework , which provides a self-supervised solution to shadow removal with unpaired data. Specifically, our framework comprises two major stages: coarse processing stage and refined restoration stage, which correspond to the semantic-guided generative adversarial network (SG-GAN) and the diffusion-based restoration module (DBRM) respectively. In the first stage, SG-GAN, consists of two sub-branches. We train mask-guided generators to synthesize shadow images, assisting the shadow removal network in adversarial training. Then, we propose a multi-modal semantic prompter (MSP) that uses pre-trained visual-language models to extract features and semantic information from real images and text, enhancing shadow removal performance in both branches. In the second stage, we use paired data as the input which is constructed by the cycle-consistent structure in the branch of SG-GAN. We exploit DBRM to further refine the coarse results obtained in SG-GAN, which might contain edge artifacts and texture blurring, making them closer to the target images. This process overcomes the obstacle that the diffusion model relies on paired data for training, which otherwise makes its use in unsupervised methods challenging.

In summary, our main contributions of this work are as follows:
\begin{itemize}
    \item We propose a semantic-guided adversarial diffusion model for self-supervised shadow removal to solve the difficulty of diffusion model processing unpaired data by constructing paired images with a cycle-consistent structure. Our methods can learn to remove shadow from unpaired data and solve the problem of obvious shadow boundary and texture detail missing in results.
    \item A general-purpose multi-modal semantic prompter is introduced to bridge the inherent gap between real-world shadow images and synthetic shadow images. Meanwhile, the effectiveness of this module is validated in other methods.
    \item We conduct extensive experiments on two public datasets. The experimental results show that our proposed method achieves competitive performance and is superior to previous unsupervised shadow removal methods.
\end{itemize}

%-------------------------------------------------------------------------
\section{Related work}

%-------------------------------------------------------------------------
\subsection{Shadow removal}

Traditional shadow removal methods rely on image gradients\cite{2006On}, illumination information\cite{2013Fast}, and image intensity regions\cite{2013Paired} to remove shadows. These early shadow removal methods often model the image without shadows or transfer color and texture features from non-shadow regions to shadow regions to achieve shadow removal. However, due to the lack of accuracy in underlying physical models, these methods usually cannot handle shadows in complex real-world scenes. With the emergence of deep learning methods, deep learning-based approaches\cite{yang2021corporate}\cite{yang2022domfn}\cite{yang2022exploiting}\cite{fu2024noise} have shown greater advantages in dealing with more complex and varied scenes.

Qu\cite{2017DeshadowNet} proposed an end-to-end deep neural network DeshadowNet for shadow removal, which predicts output shadows from three different directional information using multiple contextual architectures. Wang\cite{2018Stacked} designed ST-CGAN to detect and remove shadows and created the first large-scale shadow benchmark dataset ISTD with three pairs of 1870 paired images. Due to the difficulty and inconsistency in obtaining paired shadow images in practice, to eliminate the reliance on paired data, Hu\cite{2019Mask} proposed the Mask-ShadowGAN method based on the idea of CycleGAN, treating the shadow removal problem as an image-to-image style transfer problem. LG-ShadowNet\cite{liu2021Lg} proposed a shadow image enhancement method  based on a simple physical lighting model and image decomposition formula for shadow and pseudo-shadow removal. Liu\cite{liu2021shadow} introduced G2R-ShadowNet for shadow removal using a training dataset constructed from a set of shadow images and their corresponding shadow masks. Most of these methods adopt adversarial learning, which results in obvious shadow boundaries and lack of texture details.
\begin{figure*}[tbp]
  \centering
  \mbox{} \hfill
  % the following command controls the width of the embedded PS file
  % (relative to the width of the current column)
  \includegraphics[width=.9\linewidth]{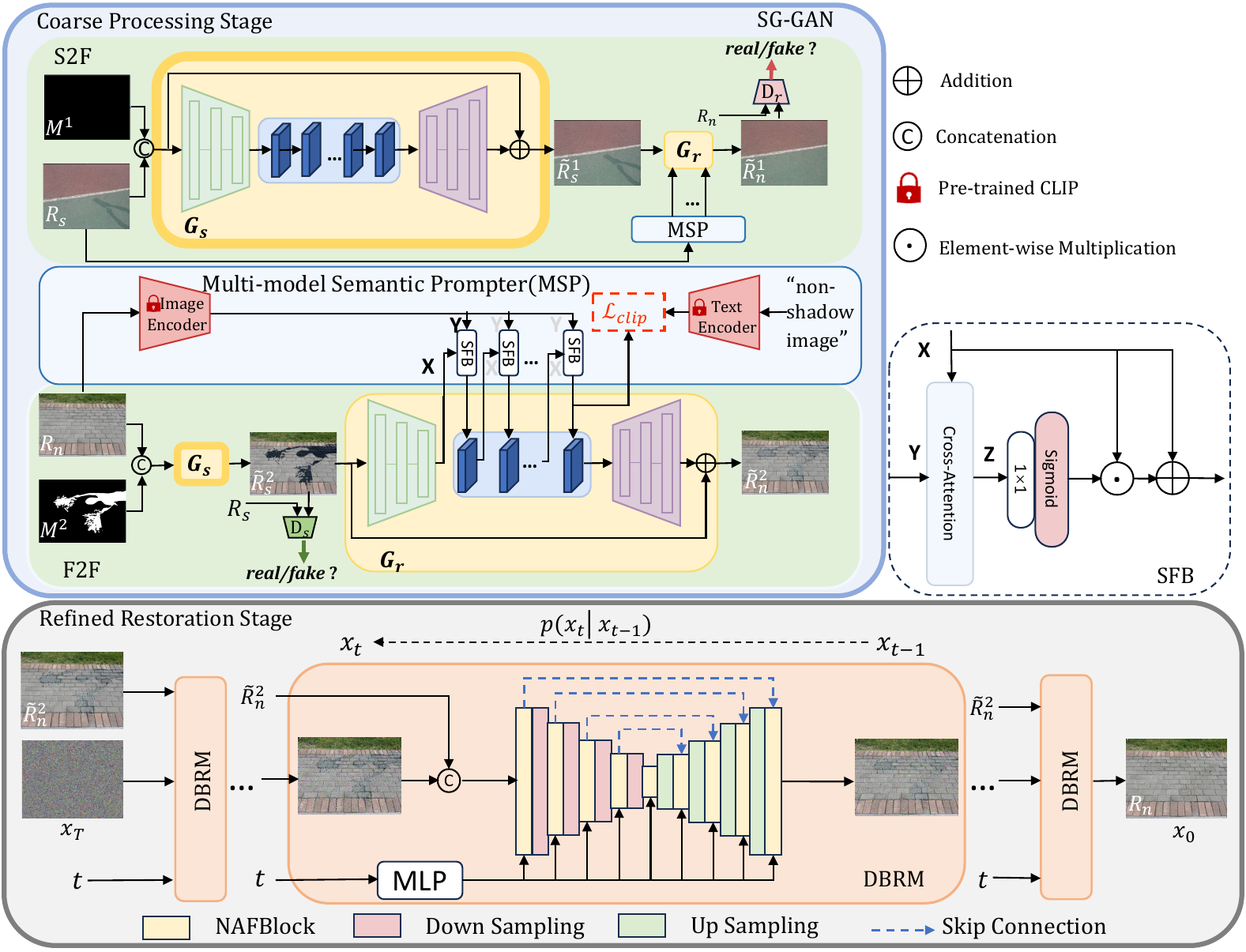}
  % replacing the above command with the one below will explicitly set
  % the bounding box of the PS figure to the rectangle (xl,yl),(xh,yh).
  % It will also prevent LaTeX from reading the PS file to determine
  % the bounding box (i.e., it will speed up the compilation process)
  % \includegraphics[width=.3\linewidth, bb=39 696 126 756]{sampleFig}
  \hfill
  \hfill \mbox{}
  \caption{\label{fig:ex2}%
           Overall pipeline of our method. At the coarse processing stage, the SG-GAN, which consists of S2F, F2F, and MSP, predicts the coarse shadow removal results ${\widetilde{R}}^2_s$. At the refined restoration stage, DBRM takes paired data $R_n$ and ${\widetilde{R}}^2_s$ from the previous stage as input, where the coarse result ${\widetilde{R}}^2_s$ is refined. }
\end{figure*}

%-------------------------------------------------------------------------
\subsection{Diffusion model}
Diffusion Models\cite{sohl2015deep} \cite{ho2020denoising} are generative models which learn distributions of real images by the Gaussian noise blurring process and the reverse denoising process. They have been successfully applied to various computer vision tasks, e.g., image super-resolution \cite{saharia2022image}, inpainting \cite{lugmayr2022repaint}, color harmonization \cite{xu2023learning} and image restoration \cite{guo2023boundary} \cite{Zhaowfdiff}. 

In recent years, with the proposed diffusion model, diffusion models have also used in shadow removal tasks. For example, \cite{mei2024latent} enhanced the diffusion process by conditioning on a learned latent feature space from shadow-free images, while integrating noise features to avoid local optima during training. Liu\cite{liu2024recasting} used the diffusion model to reconstruct the local illumination of the shadow region under the condition of the global illumination of the shadow image. However, these methods using the diffusion model inevitably need to use paired data to provide supervisory information for network training. 

Additionally, methods combining adversarial learning with diffusion models have emerged. \cite{kim2022diffusion} employed DDPM with adversarial learning for unsupervised vessel segmentation and has achieved promising results. However, as far as we know, no study have yet combined these approaches in the domain of shadow removal.

%-------------------------------------------------------------------------
\section{Preliminary}
In this section, we briefly review the key concepts underlying SDE-based diffusion models and outline the process of generating samples with reverse-time SDEs. Let $p_0$ denote the initial data distribution, and $t \in [0, T]$ denote the continuous time variable. We consider a diffusion process ${x(t)}_{t=0}^{T}$ defined by an SDE of the form:
\begin{equation}
    dx = f(x, t)dt + g(t) dw, \quad x(0) \sim p_0(x),
\end{equation}
where $f$ and $g$ are the drift and dispersion functions, respectively, $w$ is a standard Wiener process, and $x(0) \in \mathbb{R}^d$ is the initial condition. Typically, the terminal state $x(T)$ follows a Gaussian distribution with fixed mean and variance. The idea is to design such an SDE that gradually transforms the data distribution into Gaussian noise\cite{de2022riemannian, lu2022dpm}.

We can reverse the process to sample data from noise by simulating the SDE backward in time\cite{song2020score}. \cite{anderson1982reverse} shows that a reverse-time representation of the SDE (1) is given by:
\begin{equation}
    dx = \left[ f(x, t) - g(t)^2 \nabla_x \log p_t(x) \right] dt + g(t) d\hat{w},
\end{equation}
where $x(T) \sim p_T(x)$. Here, $\hat{w}$ is a reverse-time Wiener process, and $p_t(x)$ is the marginal probability density function of $x(t)$ at time $t$. Since the score function $\nabla_x \log p_t(x)$ is generally intractable, SDE-based diffusion models approximate it by training a time-dependent neural network $s_\theta(x, t)$ using a score matching objective\cite{song2020score}.
%------------------------------------------------------------------------
\section{Proposed method}
\begin{figure}[htb]
  \centering
  % the following command controls the width of the embedded PS file
  % (relative to the width of the current column)
  \includegraphics[width=1\linewidth]{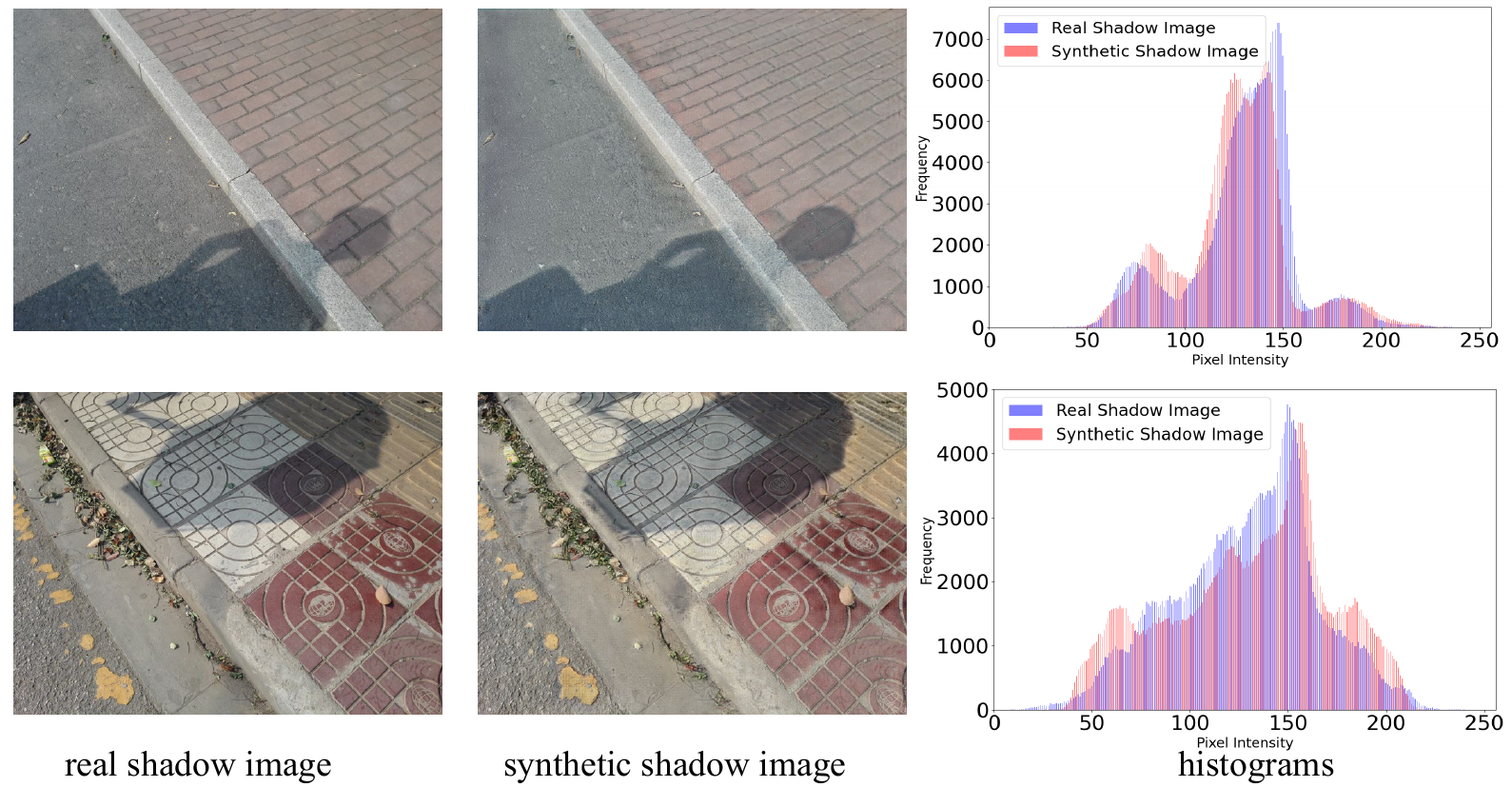}
  % replacing the above command with the one below will explicitly set
  % the bounding box of the PS figure to the rectangle (xl,yl),(xh,yh).
  % It will also prevent LaTeX from reading the PS file to determine
  % the bounding box (i.e., it will speed up the compilation process)
  % \includegraphics[width=.95\linewidth, bb=39 696 126 756]{sampleFig}
  %
  \caption{\label{fig:histograms}
            Real shadow images from ISTD and corresponding synthetic shadow images obtained from our generator $G_s$, the histograms show the inconsistencies in the intensity distribution between them.}
\end{figure}
%-------------------------------------------------------------------------
\subsection{ Overall framework}

To enhance the effectiveness of shadow removal networks and address edge artifacts and  blur texture in unsupervised methods, we propose a semantic-guided adversarial diffusion model for self-supervised shadow removal. Figure~\ref{fig:ex2} illustrates the overall network architecture of our method, which consists of two stages: coarse processing stage and refined restoration stage.These two stages are composed of the semantic-guided generative adversarial network (SG-GAN) and the diffusion-based restoration module (DBRM) respectively. In SG-GAN, we use a general-purpose multi-modal semantic prompter (MSP) to extract semantic information by the pre-trained clip model helps the network in better restoration. In terms of the structure of SG-GAN, it can be divided into two branches which utilizes a set of unpaired shadow images, shadow-free images, and shadow masks as inputs for training. It includes shadow generation and removal generators \(G_s\) and \(G_r\), as well as shadow and shadow-free image discriminators \(D_s\) and \(D_r\). However, SG-GAN relies on discriminators and consistency constraints, which are insufficient to achieve optimal results. The outcomes after shadow removal still contain undesirable noise.Therefore, in refined restoration stage, DBRM uses the coarse shadow removal results obtained from SG-GAN to construct paired data with the clean inputs for network training. By the powerful generative capabilities of the diffusion model, DBRM further refines the coarse results to remove artifacts and enhance texture details.

\subsection{Multi-modal semantic prompter} Existing unsupervised methods \cite{liu2021shadow} \cite{2019Mask} primarily rely on generated shadow images (synthetic data) to train shadow removal networks without the guidance from real prior information. However, as shown in Figure~\ref{fig:histograms}, even though the real shadow image and the synthesized shadow image look very similar, data distribution difference still exists in. The shadow removal network trained with synthetic images might reduce its effectiveness on real-world shadow images.

Useful prompts can help to correct task networks toward better performance. To minimize the impact of this gap, we designed a multi-modal semantic prompter (MSP). As shown in Figure~\ref{fig:ex2}, MSP extracts features using the image encoder $C_{image}$ and text encoder $C_{text}$ from a pre-trained CLIP model \cite{radford2021learning}. The prior information extracted by the image encoder is fused with the features extracted by $G_r$ in residual blocks through semantic fusion blocks (SFB), while the text features extracted by the text encoder are used to define a clip loss (introduced in Sec.\ref{sec:4.3}) to further constrain the image recovered by $G_r$.

SFB aims to better perceive prior information, using this more reliably perceived content to assist $G_r$ in restoration. It also controls the propagation of perceived prior information, enabling it to adaptively learn more useful features in $G_r$ for better restoration. Given a recovery tensor $X_{k-1}$ from the $(k-1)_{th}$ residual block in $G_r$ and a semantic tensor $Y $ extracted by $C_{image}$, they are first fused through a cross attention mechanism to obtain $Z_{k-1}$:
\begin{equation}
    Z_{k-1} = A_{c-attn}(X_{k-1},Y)
\end{equation}
where $A_{c-attn}$ refers to the cross attention.
Then, using a $1 \times1$ convolution and a gated control function sigmoid, the input for the next residual block and can be obtained:
\begin{equation}
    \mathcal{S}(X_{k-1}, Y) = \sigma(W_p(Z_{k-1})) \odot X_{k-1} + X_{k-1}
\end{equation}
where $\sigma(\cdot)$ represents the sigmoid function and $W_p(\cdot)$ means the 1 × 1 point-wise convolution.

%-------------------------------------------------------------------------
\subsection{Semantic-guided generative adversarial network}
\label{sec:4.3}
SG-GAN consists of two branches: shadow to shadow-free (S2F) and shadow-free to shadow-free (F2F), of which the input is a real-world image and a shadow mask. The input shadow mask is a binary map where 0 represents non-shadow (black) regions and 1 represents shadow (white) regions. 

\noindent\textbf{S2F and F2F sub-branches} 
In S2F, a shadow mask image $M^1$ consisting entirely of zeros and a real shadow image $R_{s}$ are used as inputs to the generator $G_{s}$ to produce an image ${\widetilde{R}}^1_s $ without additional shadows:
\begin{equation}
	{\widetilde{R}}^1_s=G_s(R_s,M^1)
\end{equation}
The generated shadow image ${\widetilde{R}}^1_s$ is then transformed into a shadow-free image ${\widetilde{R}}^1_n$ using the generator $G_r$. Subsequently, the discriminator $ D_r$ is employed to distinguish whether ${\widetilde{R}}_n$ is a genuine shadow-free image:
\begin{equation}
	{\widetilde{R}}^1_n=G_r({\widetilde{R}}^1_s), D_r({\widetilde{R}}^1_n)=real/fake?
\end{equation}

In F2F,  a real shadow-free image $R_n$ and a shadow mask $M^2$ which indicates the shadow region are required as input. We use the method described in \cite{liu2021shadow} to generate the shadow mask $M^2$. Then, with above inputs, generator $G_s$ generates a shadow image ${\widetilde{R}}^2_s $ to deceive the discriminator $D_s $, making it difficult for $D_s$ to discern whether it is a genuine shadow image:
\begin{equation}	{\widetilde{R}}^2_s=G_s(Rn,M^2), D_s({\widetilde{R}}^2_s)=real/fake?
\end{equation}

Then the synthesized ${\widetilde{R}}^2_s $ as input to generator $G_r$ for shadow removal to obtain a coarse shadow-free image ${\widetilde{R}}^2_n $.

In the above process, we find that the input to generator $G_r$ is always the synthesized shadow image generated by $G_s$. In order to improve the shadow removal performance of $G_r$, we add MSP in $G_r$ to reduce the impact of synthesized data in both S2F and F2F.

The architectures of $G_r$ and $G_r$ are same, which mainly follow the generator proposed by Hu \cite{2019Mask}. They consist of three convolution layers with a stride of 2, followed by nine residual blocks for feature extraction, and finally three deconvolution layers to upsample feature map. Instance normalization is applied after each convolution operation. For the discriminators $D_r$ and $D_s$ we adopt the architecture proposed in PatchGAN \cite{isola2017image}.
%-------------------------------------------------------------------------

\noindent\textbf{Loss function} In S2F, we use identity loss to make the generated shadow image ${\widetilde{R}}^1_s $ close to the input shadow image $R_s $ :
\begin{equation}
 \mathcal{L}_{\textit{identity}}(G_s) = \mathbb{E}_{R_s \sim p(R_s)} \left[ \| G_s(R_s, M^1) - R_s \|_1 \right] 
\end{equation}

For generator $G_r $ and its discriminator$ D_r$, the objective function is optimized as follows:
\begin{equation}
\begin{aligned}
\mathcal{L}_{\textit{GAN}_r} &= \mathbb{E}_{R_n \sim p(R_n)} \left[ \log(D_r(R_n)) \right] \\
&\quad + \mathbb{E}_{R_s \sim p(R_s)} \left[ \log(1 - D_r(G_r(\widetilde{R}^1_s))) \right]
\end{aligned}
\end{equation}

We incorporated the MSP into $G_r$. This process introduces a clip loss to help $G_r$ better remove shadows. As shown in Figure~\ref{fig:ex2}, the clip loss constraint between the output $\mathcal{S}(X, C_{image}(R_s)$ of the last SFB and the semantic features extracted from input text $T$ by clip text encoder $C_{text}$ is defined as follows:
\begin{equation}
\mathcal{L}_{\textit{clip}_{S2F}} =  \frac{e^{cos\left(\mathcal{S}(X, C_{image}(R_s),C_{text}(T))\right)/\tau}}{e^{cos\left(\mathcal{S}(X, C_{image}(R_s),C_{text}(T)\right)/\tau}+e^{\frac{1}{\tau}}}
\end{equation}
where $X$ is the output of the penultimate residual block in $G_r$, $\tau$ denotes the temperature parameter which we set to 0.5 in experiment.
In F2F, the adversarial loss for generator $G_s$ and discriminator $D_s$ is formulated as:
\begin{equation}
\begin{aligned}
\mathcal{L}_{\text{GAN}_s} &= \mathbb{E}_{R_s \sim p(R_s)} \left[ \log(D_s(R_s)) \right] \\
&+ \mathbb{E}_{R_n \sim p(R_n)} \left[ \log(1 - D_s(G_s(R_n, M^2))) \right]
\end{aligned}
\end{equation}

MSP is also used in F2F, so we define a loss similair to $\mathcal{L}_{\textit{clip}_{S2F}}$ that makes the shadow removal result close to the input text $T$:
\begin{equation}
\mathcal{L}_{\textit{clip}_{F2F}} =  \frac{e^{cos\left(\mathcal{S}(X, C_{image}(R_s),C_{text}(T)\right)/t}}{e^{cos\left(\mathcal{S}(X, C_{image}(R_s),C_{text}(T))\right)/t}+e^{\frac{1}{t}}}
\end{equation}

To ensure that ${\widetilde{R}^2}_n $ is similar to the original input real shadow-free image $R_n$,  we use cycle consistency loss to optimize the mapping functions in $G_s$ and $G_r$:
\begin{equation}
\mathcal{L}_{\textit{cycle}}(G_s, G_r) = \mathbb{E}_{R_n \sim p(R_n)} \left[ \| G_r(G_s(R_n, M^2)) - R_n \|_1 \right]
\end{equation}

To further emphasize that the shadow region in ${\widetilde{R}}^2_n $ guided by the shadow mask matches the content in the input image $R_n$, we apply a shadow loss as follow:
\begin{equation}
\mathcal{L}_{\textit{shadow}}(G_s, G_r) = \mathbb{E}_{R_n \sim p(R_n)} \left[ \| M \odot (G_r(G_s(R_n, M^2)) - R_n) \|_1 \right]
\end{equation}

In summary, the total loss in SG-GAN is defined as:
\begin{equation}
\begin{aligned}
\mathcal{L} = &\ \mathrm{\lambda}_1\mathcal{L}_{\text{identity}} +  \mathrm{\lambda}_2(\mathcal{L}_{\text{GAN}_s} + \mathcal{L}_{\text{GAN}_r}) \\
&\ + \mathrm{\lambda}_3(\mathcal{L}_{\textit{clip}_{S2F}} + \mathcal{L}_{\textit{clip}_{F2F}})+ \mathrm{\lambda}_4(\mathcal{L}_{\text{cycle}}+ \mathcal{L}_{\text{shadow}})
\end{aligned}
\end{equation}
where $\lambda_1$, $\lambda_2$, $\lambda_3$, and $\lambda_4$ are weights balancing different loss terms. In our experiment, we set $\lambda_1$, $\lambda_2$, $\lambda_3$ , and $\lambda_4$ to 5, 1, 0.2, and 10, respectively.
%-------------------------------------------------------------------------
\subsection{ Diffusion-based restoration module}
\begin{figure*}[tbp]
  \centering
  \mbox{} \hfill
  % the following command controls the width of the embedded PS file
  % (relative to the width of the current column)
  \includegraphics[width=1\linewidth]{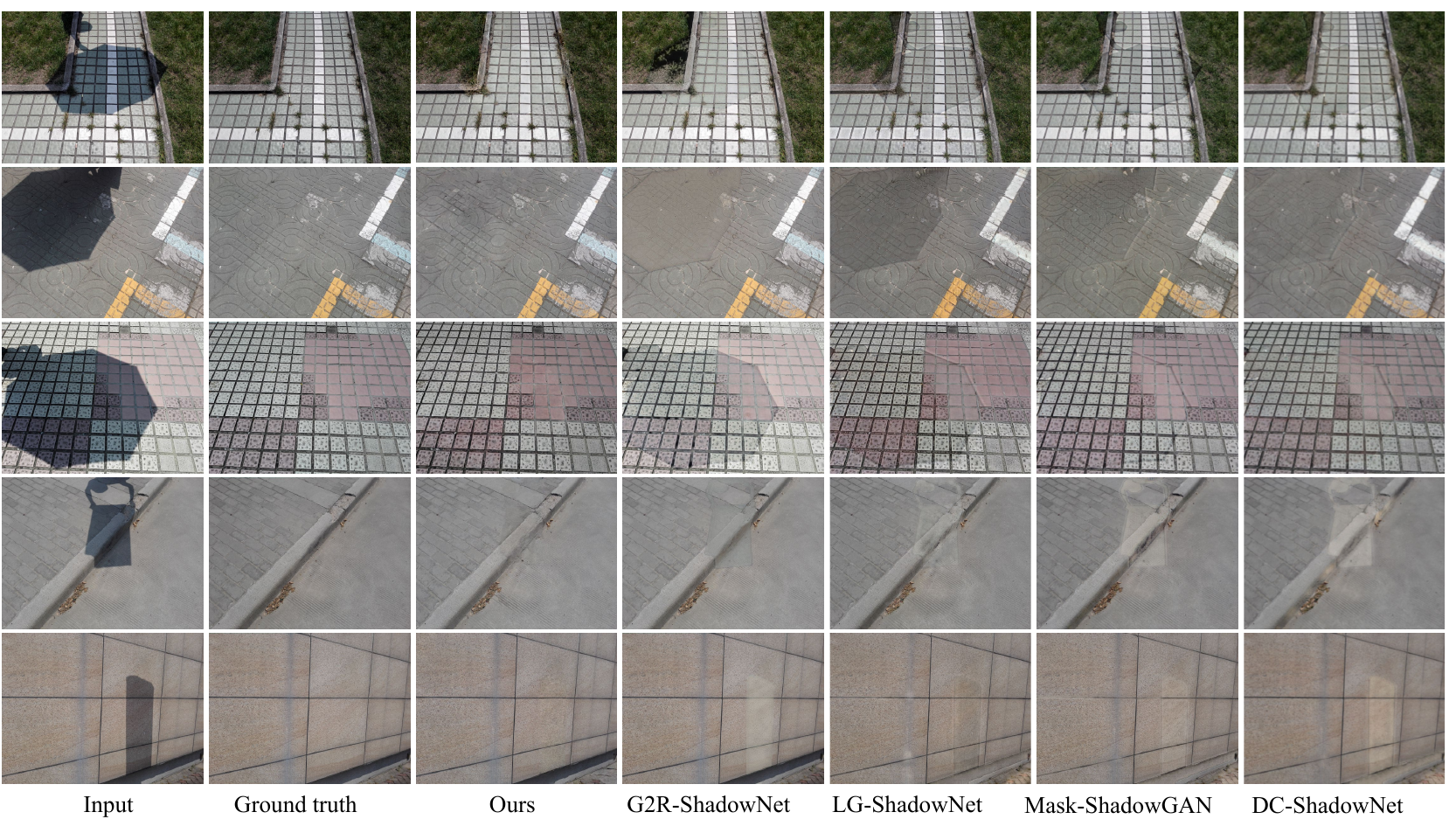}
  % replacing the above command with the one below will explicitly set
  % the bounding box of the PS figure to the rectangle (xl,yl),(xh,yh).
  % It will also prevent LaTeX from reading the PS file to determine
  % the bounding box (i.e., it will speed up the compilation process)
  % \includegraphics[width=.3\linewidth, bb=39 696 126 756]{sampleFig}
  \hfill
  \caption{\label{fig:ex4}%
           Visualisation comparisons results on five real-world challenging samples from the ISTD dataset.  }
\end{figure*}
At this stage, we employ IR-SDE \cite{luo2023image} as the diffusion framework for diffusion model. It allows us to better understand and control the image generation process by simulating the image degradation process. The key idea of our diffusion model is to combine a mean-reverting SDE with a maximum likelihood objective for neural network training, which naturally transforms high-quality images into degraded low-quality images regardless of the complexity of the degradation. 

\noindent\textbf{Diffusion framework} 
According to \cite{luo2023image}, the forward process of a mean-reverting SDE is defined as:
\begin{equation}
    dx_t = \theta_t(\mu - x_t) dt + \sigma_t dw,
\end{equation}
where $\theta_t$ and $\sigma_t$ are time-varying positive parameters representing the mean regression rate and stochastic volatility, respectively. Here, $\mu$ is the state mean, and $w$ denotes Brownian motion.

In our model, a coarse shadow removal result ${\widetilde{R}}^2_n$ is obtained in the second branch of SG-GAN. We use ${\widetilde{R}}^2_n$ and the corresponding input clean image $R_n$ to construct paired data (LQ and GT) for our DBRM. Given paired images, we set $R_n$ as the initial state $x_0$ and ${\widetilde{R}}^2_n$ as $\mu$, with a fixed noise level $\lambda$.

By ensuring $\sigma_t^2 / \theta_t = 2 \lambda^2$ for all $t$, the forward process solution is:
\begin{equation}
x(t) = {\widetilde{R}}^2_n + (x(s) - {\widetilde{R}}^2_n) e^{-\bar{\theta}{s:t
}} + \int{s}^{t} \sigma_z e^{-\bar{\theta}{z:t}}dw(z),
\end{equation}
where $\bar{\theta}{s:t} = \int_s^t \theta_z , dz$. The transition kernel is:
\begin{equation}
p(x(t) \mid x(s)) = \mathcal{N}(x(t) \mid m_{s:t}(x(s)), v_{s:t}),
\end{equation}
where $m_{s:t}$ is the mean and $v_{s:t}$ is the variance. The forward SDE iteratively transforms the GT image $R_n$ into the LQ image $\widetilde{R}^2_n$ with added noise:
\begin{equation}
p(x_t \mid x_{t-1}) = \mathcal{N} \left( x_t \mid m_{t-1}(x_{t-1}), v_{t-1
} \right).
\end{equation}
A notable property of this process is that noisy data $x_t$ can be sampled from $x_0$ in closed form:
\begin{equation}
    p_t(x) = \mathcal{N}\left(x(t) \mid m_t(x), v_t\right)  
\end{equation}
with $m_t = \widetilde{R}^2_n + (x(0) - \widetilde{R}^2_n) e^{-\bar{\theta} t}$, $v_t = \lambda^2 \left(1 - e^{-2 \bar{\theta} t}\right)$ and $\bar\theta_t$ referring to $m_{0:t}$, $v_{0:t}$ and $\bar\theta_{0:t}$, respectively.

The reverse-time representation from \cite{anderson1982reverse} is:
\begin{equation}
dx = [\theta_t({\widetilde{R}}^2_n - x) - \sigma_t^2 \nabla_x \log p_t(x)] dt + \sigma_t dw,
\end{equation}
where $\nabla_x \log p_t(x)$ is the score of the marginal distribution at time $t$. Given the GT image $R_n$, we compute the score function as:
\begin{equation}
    \nabla_x \log p_t(x) = -\frac{x(t) - m_t}{v_t}.
\end{equation}

We then train a conditional time-dependent neural network $\tilde{\epsilon}_{\phi}(x_t, \widetilde{R}^2_n, t)$ to estimate noise.  Sampling $x_t$ according to $ x_t = m_t(x) + \sqrt{v_t}\ \epsilon_t$, where $\epsilon_t \sim \mathcal{N}(0, I)$. The score can then be directly computed from the noise:
\begin{equation}
\nabla_x \log p_t(x) = -\frac{\epsilon_t}{\sqrt{v_t}}.
\end{equation}

\begin{figure*}[tbp]
  \centering
  \mbox{} \hfill
  % the following command controls the width of the embedded PS file
  % (relative to the width of the current column)
  \includegraphics[width=1\linewidth]{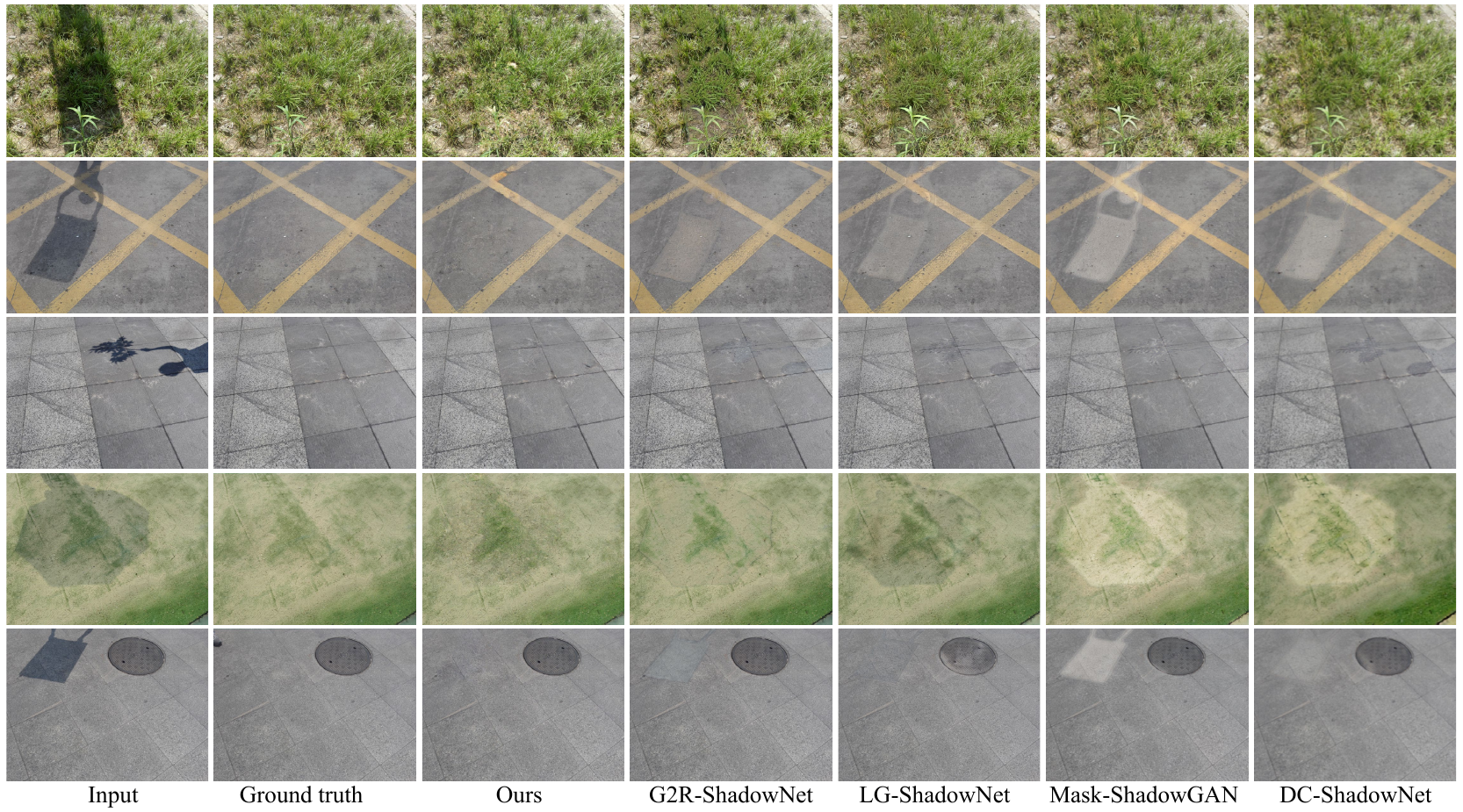}
  % replacing the above command with the one below will explicitly set
  % the bounding box of the PS figure to the rectangle (xl,yl),(xh,yh).
  % It will also prevent LaTeX from reading the PS file to determine
  % the bounding box (i.e., it will speed up the compilation process)
  % \includegraphics[width=.3\linewidth, bb=39 696 126 756]{sampleFig}
  \hfill
  \caption{\label{fig:ex4-2}%
           Visualisation comparisons results on five real-world challenging samples from the AISTD dataset.  }
\end{figure*}

\noindent\textbf{Network architecture} As shown the architecture of DBRM in Figure~\ref{fig:ex2}, our noise prediction network is based on nonlinear activated free block (NAF block) \cite{chen2022simple}. NAF block replaces nonlinear activations (such as ReLU and GELU) with a SimpleGate unit. Given an input, SimpleGate splits it into two features along the channel dimension and then uses a linear gate to compute the output. The SimpleGate unit is added after the depth-wise convolution and between the two fully connected layers. Meanwhile, we add an additional multi-layer perceptual processing time embedding for each NAF block.

%-------------------------------------------------------------------------

\noindent\textbf{Loss function}  An alternative maximum likelihood objective aims to find the optimal trajectory $x_{1:T}$ given the high-quality image $x_0$, stabilizing training and recovering more accurate images. Following IR-SDE \cite{luo2023image}, we train our prediction network with a maximum likelihood loss which specifies the optimal reverse path $ x^*_{t-1}$ for all times: 
\begin{equation}
    \begin{aligned}
x^*_{t-1} = & \frac{1 - e^{-2 \bar{\theta}_{t-1}}}{1 - e^{-2 \bar{\theta}_t}} e^{-\theta'_t} (x_t - \widetilde{R}^2_n) \\
& + \frac{1 - e^{-2 \theta'_t}}{1 - e^{-2 \bar{\theta}_t}} e^{-\bar{\theta}_{t-1}} (R_n - \widetilde{R}^2_n) + \widetilde{R}^2_n
\end{aligned}
\end{equation}
where $\theta'_i = \int_{i-1}^{i} \theta_t \, dt$. Then, we define the maximum likelihood loss :
\begin{equation}
    \mathcal{L}_{\textit{diff}}= \sum_{t=1}^{T} \gamma_t \mathbb{E} \left[ \left\| x_t - (dx_t) \tilde{\epsilon}_{\phi}-{x^*_{t-1}} \right\| \right]
\end{equation}
where $\gamma_1, . . . , \gamma_T $ are positive weights and ${\{x_t\}}^T_{t=0} $ denotes the discretization of the diffusion process. $(dx) \tilde{\epsilon}_{\phi}$ denotes the reverse-time SDE in Eq.17 and its score is predicted by the noise network $\tilde{\epsilon}_{\phi}$. $x_t - (dx_t) \tilde{\epsilon}_{\phi}$ is the reverse $x_{t-1}$. Once trained, we can use the network $ \tilde{\epsilon}_{\phi}$ to generate high-quality images by sampling a noisy state $x_T$ and iteratively solving the Eq.17 with a numerical scheme.

%-------------------------------------------------------------------------
\section{Experiment}
 \begin{table*}
	\begin{center}
    \mbox{} \hfill
	\resizebox{1.0\textwidth}{!}{
	\begin{tabular}{|c|c|ccc|ccc|cccc|} % 指定表格列数和对齐方式
		\hline % 顶部横线
		\multirow{2}*{Scheme} & \multirow{2}*{Methods} & \multicolumn{3}{c|}{Shadow Region} & \multicolumn{3}{c|}{Non-shadow Region} & \multicolumn{4}{c|}{All Image}\\ 		
		\cline{3-12}
		& & RMSE\(  \downarrow \) & PSNR\(  \uparrow \) & SSIM\(  \uparrow \) &  RMSE\(  \downarrow \) & PSNR\(  \uparrow \) & SSIM\(  \uparrow \)& RMSE\(  \downarrow \) & PSNR\(  \uparrow \) & SSIM\(  \uparrow \) & LPIPS\(  \downarrow \)  \\
		\hline % 中间横线
		\multirow{4}*{Supervised} & DHAN &7.53 &\underline{35.8}2 & \underline{0.989} &5.33 &30.95 &0.971 &5.68 &29.09 & \underline{0.953} & \textbf{0.027}\\ % 表格内容
		& Sp+M-Net &\underline{7.13}& 35.08 &0.984 &\textbf{3.16} &\textbf{36.38} &\textbf{0.979} &\textbf{3.92} &\underline{31.89} &\underline{0.953} & 0.079 \\
		& AEFNet &7.91 &34.71 &0.975 &5.51 &28.61 &0.880 &5.88 &27.19 & 0.945 & \underline{0.045} \\
        & ShadowDiffusion &\textbf{4.10} & \textbf{40.06} & \textbf{0.996} & \underline{4.18} &\underline{33.00} & \underline{0.973} &\underline{4.16 }&\textbf{32.20} &\textbf{0.967} & -\\
		\hline
		\multirow{5}*{Unsupervised} & Mask-ShadowGAN & 13.00 & 30.53 & \underline{0.977} & \underline{6.07} & \underline{28.86} & 0.960 & 6.96 &25.72 & 0.925 & 0.064 \\
		& DC-ShadowNet & 11.89 & 31.27 & 0.969 & 7.84 & 27.20 & 0.910 & 7.03 & 25.51 & 0.865 & 0.104 \\
		& LG-ShadowNet & \underline{10.72} & \underline{31.53} & \textbf{0.979} & 6.30 & 27.67 & \underline{0.967} & \textbf{6.08} &\underline{26.62} & \textbf{0.936} & \underline{0.056} \\
            &S3R-Net & 12.16 &- &- &6.38 &- &- & 7.12&- &- &- \\
		& Ours & \textbf{10.53} & \textbf{32.68} & 0.966 & \textbf{5.54} & \textbf{30.96} & \textbf{0.970} & \underline{6.29} & \textbf{27.94} & \underline{0.930} & \textbf{0.038} \\
		\hline % 底部横线
		& Shadow image& 32.10 & 22.40 & 0.936 & 7.09 & 27.32 &0.976 &10.89 & 20.56 & 0.893 & 0.116 \\ % 表格内容
		\hline % 底部横线
	\end{tabular}
	}
 \hfill
	\end{center}
	\caption{\label{tab:tab1}%
        Quantitative comparison results of our methods with the state-of-the-art methods on ISTD dataset. The best and second performances for supervised learning and unsupervised learning methods are highlighted in \textbf{Bold} and
underlined, respectively. ‘-’ denotes the results are not publicly available.}
\end{table*}
\begin{table*}
	\begin{center}
	\resizebox{1.0\textwidth}{!}{
	\begin{tabular}{|c|c|ccc|ccc|cccc|} % 指定表格列数和对齐方式
		\hline % 顶部横线
		\multirow{2}*{Scheme} & \multirow{2}*{Methods} & \multicolumn{3}{c|}{Shadow Region} & \multicolumn{3}{c|}{Non-shadow Region} & \multicolumn{4}{c|}{All}\\ 		
		\cline{3-12}
		& & RMSE\(  \downarrow \) & PSNR\(  \uparrow \) & SSIM\(  \uparrow \) &  RMSE\(  \downarrow \) & PSNR\(  \uparrow \) & SSIM\(  \uparrow \)& RMSE\(  \downarrow \) & PSNR\(  \uparrow \) & SSIM\(  \uparrow \) & LPIPS\(  \downarrow \)  \\
		\hline % 中间横线
		\multirow{4}*{Supervised} & DHAN & 9.57 & 32.92 & 0.987 & 7.41 & 27.15 &0.972  & 7.77 &25.66 &0.954 & \textbf{0.026}\\ % 表格内容
		& Sp+M-Net & \underline{5.91} & \underline{37.60} & \underline{0.99} 0& \underline{2.99} & \underline{36.02} & \underline{0.976} & \underline{3.46} &\underline{32.94} & \underline{0.962} & 0.085\\
		& AEFNet & 6.55 & 36.04 & 0.978& 3.77 & 31.16 &0.892 & 4.22 &29.45 &0.861 & \underline{0.046} \\
        &ShadowDiffusion &\textbf{4.60} & \textbf{40.13} & \textbf{0.997} &\textbf{2.74} & \textbf{36.48}& \textbf{0.979} &\textbf{2.91} & \textbf{35.66} & \textbf{0.974} & -\\
        \hline
		\multirow{5}*{Unsupervised} & Mask-ShadowGAN & 11.28 &  31.50& \underline{0.981} & 3.90 & 32.63  & 0.967 &  4.97&  28.11& 0.936 & 0.063\\ % 表格内容
		& DC-ShadowNet & 10.81 & 32.15 & 0.978&  3.46 & \underline{35.50} &0.974 &4.61& 29.09& \underline{0.940} & 0.051\\
		& LG-ShadowNet & \underline{9.90} & \underline{32.42} & \textbf{0.982}& \underline{3.18} & 34.01& \underline{0.976}& \underline{4.25} & \underline{29.31} & \textbf{0.947} & \underline{0.049} \\
            & S3R-Net & 12.86 &- & -& 4.43 &- & -&  5.71 &- &- & -\\
         & Ours & \textbf{9.48} & \textbf{33.63} & 0.968 & \textbf{3.09} & \textbf{36.01} & \textbf{0.978} & \textbf{4.06} & \textbf{31.09} & \underline{0.940} &  \textbf{0.034}\\
		\hline % 底部横线
		& Shadow image& 36.95 & 20.83 & 0.927 & 2.42 & 37.46 &0.985 &8.40 &20.46 &0.894 & 0.117\\ % 表格内容
		\hline % 底部横线
	\end{tabular}
    }
\end{center}
	\caption{\label{tab:tab2}Quantitative comparison results of our methods with the state-of-the-art methods on AISTD dataset.}
\end{table*}
%-----------------------------------------------------------
\subsection{ Datasets and evaluation metrics }
\textbf{Dataset} We utilize two state-of-the-art shadow removal datasets: ISTD \cite{2018Stacked} and AISTD \cite{le2019shadow}. ISTD comprising 1870 triplets of shadow, shadow-free images and shadow masks, with 1330 triplets for training and 540 triplets for testing. AISTD is the adjust dataset which further corrects the color inconsistency problem of images from the ISTD. 

\noindent\textbf{Evaluation metrics} In our experiments, we follow \cite{le2021physics} calculate root mean square error (RMSE) in the LAB color space , and employ the structure similarity (SSIM), peak signal-to-noise ratio (PSNR), and learned perceptual image patch similarity (LPIPS) as evaluation metrics for  comparisons. Generally, higher PSNR and SSIM values are preferred, while lower RMSE and LPIPS values indicate better performance. We provide the metrics measured on the shadow region, non-shadow region and whole image for reference.

%------------------------------------------------------------------------
\subsection{ Experimental settings }
We implement our methods using PyTorch \cite{paszke2019pytorch} and a single NVIDIA GeForce GTX 3090 GPU. At first stage, we initialise our SG-GAN using a Gaussian distribution with a mean of 0 and a standard deviation of 0.02. We employ the Adam optimiser to train our network with the first and the second momentum setting to 0.5 and 0.999, respectively. We train the whole model for 200 epochs and the base learning rate is set to $2 \times 10^{-4}$ for the first 100 epochs and then we apply a linear decay strategy to decrease it to 0 for the rest epochs. Additionally, horizontal flipping and random cropping strategy purposed in \cite{liu2021shadow} is applied to the training data for data augmentation. The network training involves both two branch, and they impact each other. At second stage, for training the diffusion model, we fix the noise level at 50 and set the number of diffusion denoising steps to 100. The batch sizes are set to 8 and the training patches are $256 \times 256$ pixels. We use the Lion optimizer \cite{chen2024symbolic} with $\beta_1$ = 0.9 and $\beta_2$ = 0.99. The initial learning rate is set to $3 \times 10^{-5}$ and decayed to 1e-7 by the Cosine scheduler. The noise level is fixed to 50 and the number of diffusion denoising steps is set to 100. We train DBRM for 400 000 iterations, which takes for about 4 days on the GPU. 

%------------------------------------------------------------------------
\subsection{ Comparison with the state-of-the-arts }
In this subsection, we compare our full model on the ISTD and AISTD datasets with several state-of-the-art methods, including supervised methods which are trained with paried shadow and shadow-free images: DHAN \cite{cun2020towards}, SP+M-Net \cite{le2019shadow}, AEFNet \cite{fu2021auto} and ShadowDiffusion \cite{guo2023shadowdiffusion}; unsupervised methods training without paired shadow and shadow-free images: G2R-ShadowNet \cite{liu2021shadow}, Mask-ShadowGAN \cite{2019Mask}, DC-ShadowNet \cite{jin2021dc}, LG-ShadowNet \cite{liu2021Lg} and S3R-Net \cite{kubiak2024s3r}. All of the shadow removal results by the competing methods are quoted from the original papers or reproduced using their official implementations.
\begin{figure*}[tbp]
  \centering
  \mbox{} \hfill
  % the following command controls the width of the embedded PS file
  % (relative to the width of the current column)
  \includegraphics[width=1\linewidth]{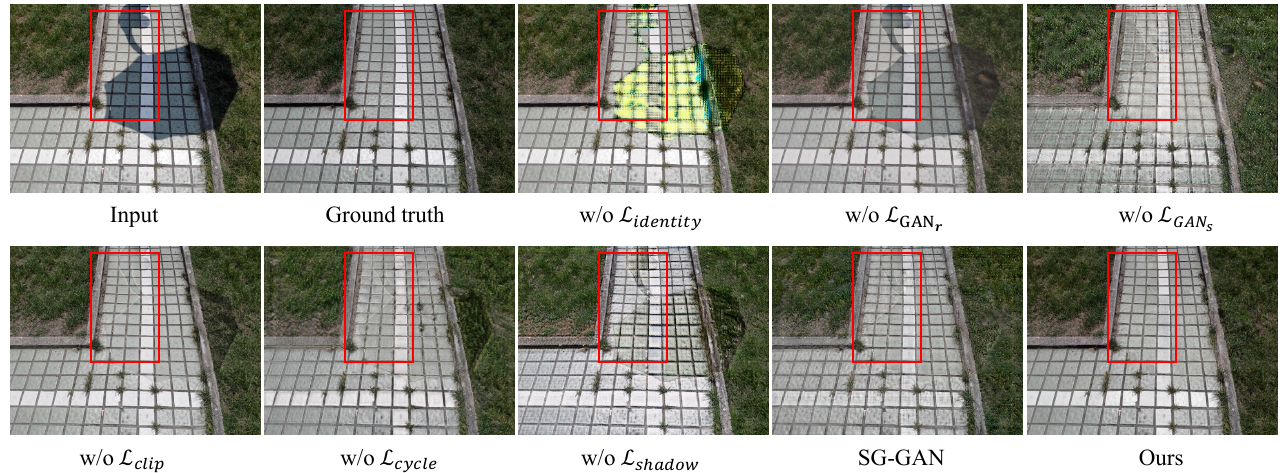}
  % replacing the above command with the one below will explicitly set
  % the bounding box of the PS figure to the rectangle (xl,yl),(xh,yh).
  % It will also prevent LaTeX from reading the PS file to determine
  % the bounding box (i.e., it will speed up the compilation process)
  % \includegraphics[width=.3\linewidth, bb=39 696 126 756]{sampleFig}
  \hfill
  \caption{\label{fig:ex5}%
           Visual comparisons result for ablation study on the use of each loss terms in SG-GAN. }
\end{figure*}

Table~\ref{tab:tab1} shows the quantitative results on the ISTD dataset. The supervised methods share the same type of training data, including shadow and shadow-free image pairs. They learn the mapping from shadow images to shadow-free images based on training pairs. However, their performance might be largely degraded when extended to some unseen scenes. Our method achieves results in shadow-free region and all image that are comparable to other deep neural networks trained on paired images, and in some metrics, it even surpasses some supervised methods. For instance, LPIPS metric of our results for the all image is better than those of Sp+M-net \cite{le2019shadow} and AEFNet \cite{fu2021auto}. Moreover, several unsupervised methods, such as  Mask-ShadowGAN \cite{2019Mask}, LG-ShadowNet \cite{liu2021Lg}, and DC-ShadowNet\cite{jin2021dc}, we can see that our method significantly outperforms these three methods. Our method outperforms LG-ShadowNet \cite{liu2021Lg} which is suboptimal on most metrics, especially the results for non-shadow region and the all image improve by about 2.1dB and 2.2dB in PSNR, respectively, except a SSIM and RMSE value is slightly lower. Additionally, our LPIPS and PSNR are clearly better than all the unsupervised methods in the table. When compared to the state-of-the-art unsupervised method S3R \cite{kubiak2024s3r}, our method performs better on RMSE. Note that the code of S3R \cite{kubiak2024s3r} is not publicly available, so we could only report their results on the dataset used in their published paper. Qualitative results on four challenging samples drawn from the testing set of ISTD are shown in Figure~\ref{fig:ex4}. Compared with other methods, our method can produce more realistic results with less artifacts and better preserve the texture details occluded by shadows. Moreover, the colour in the shadow region is more consistent with the surrounding area using our method.

We also retrain and test our model on the AISTD dataset, with quantitative results presented in Table~\ref{tab:tab2}. The comparative results are shown in Table~\ref{tab:tab2} and Figure~\ref{fig:ex4-2}. Among the unsupervised methods,  just like before our method performs the best on AISTD, followed by LG-ShadowNet \cite{liu2021Lg}. In Figure~\ref{fig:ex4-2}, the outputs generated by our competitors exhibit sharp shadow edges, whereas our results have significantly smoother shadow boundaries. Additionally, our samples show a noticeably sharper appearance in shadow region. We believe we have demonstrated that our approach can produce the most visually pleasing results to the human eye.
\begin{table}
	\begin{center}
        \resizebox{.45\textwidth}{!}{%
	\begin{tabular}{|c|c|c|c|c|}	
             \hline
			Methods & RMSE\(  \downarrow \) & PSNR\(  \uparrow \) & SSIM\(  \uparrow \) & LPIPS\(  \downarrow \) \\
            \hline
			 w/o DBRM & 7.27  & 25.44  & \textbf{0.935} & 0.054 \\	
			w/o S2F& 10.36  & 21.45 & 0.892 &0.122  \\
            w/o MSP & 7.99 & 24.89 & 0.902& 0.065 \\
			\hline
            Ours & \textbf{6.29} & \textbf{27.94} & 0.930 &  \textbf{0.038} \\
            \hline
		\end{tabular}  % 表格内容
        }
	\end{center}
	\caption{\label{tab:tab3}Ablation study on the choices of different component for our method on ISTD testing set.}
\end{table}
\begin{figure}[htb]
  \centering
  % the following command controls the width of the embedded PS file
  % (relative to the width of the current column)
  \includegraphics[width=1\linewidth]{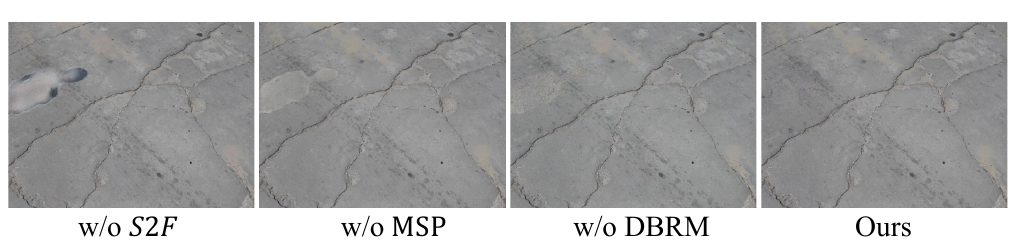}
  % replacing the above command with the one below will explicitly set
  % the bounding box of the PS figure to the rectangle (xl,yl),(xh,yh).
  % It will also prevent LaTeX from reading the PS file to determine
  % the bounding box (i.e., it will speed up the compilation process)
  % \includegraphics[width=.95\linewidth, bb=39 696 126 756]{sampleFig}
  %
  \caption{\label{fig:ex6}
             Visual comparisons result of ablation study on the use of each component for our method.}
\end{figure}
%------------------------------------------------------------------------
\subsection{ Ablation study }
To validate the efficacy of each pivotal component within our proposed method, we trained and evaluated several model variants on the ISTD dataset. First, We propose to utilize the DBRM to suppress the artifacts of shadow removal results. So we validate DBRM by removing it from our complete model, retaining only SG-GAN. Additionally, we train SG-GAN without the S2F and MSP components to evaluate their individual contributions. The quantitative results are reported in Table~\ref{tab:tab3}. Subsequently, we conduct ablation studies on loss functions we proposed. These studies involved training SG-GAN without specific loss terms to demonstrate the effectiveness of each loss function. The quantitative results are reported in Table~\ref{tab:tab4}.
\begin{table}
	\begin{center}
        \resizebox{.45\textwidth}{!}{%
	\begin{tabular}{|c|c|c|c|c|}	
             \hline
			Methods & RMSE\(  \downarrow \) & PSNR\(  \uparrow \) & SSIM\(  \uparrow \) & LPIPS\(  \downarrow \) \\
            \hline
			 w/o $\mathcal{L}_{GAN_s}$& 8.75 & 24.34 & 0.838 & 0.119 \\
			 w/o $\mathcal{L}_{GAN_r}$& 10.05 & 21.64 & 0.900 & 0.115 \\
			 w/o $\mathcal{L}_{\textit{cycle}}$& 8.40 & 24.58 & 0.825 & 0.126 \\
			 w/o $\mathcal{L}_{\textit{shadow}}$& 8.03 & 24.79 & 0.877 & 0.111 \\
			 w/o $\mathcal{L}_{\textit{identity}}$& 13.54 & 19.39 & 0.845 & 0.182 \\
             w/o $\mathcal{L}_{\textit{clip}}$& 7.41 & 25.07 & 0.927 & 0.068 \\
			\hline
            SG-GAN & \textbf{7.27} & \textbf{25.44} & \textbf{0.935} &  \textbf{0.054} \\
            \hline
		\end{tabular}  % 表格内容
        }
	\end{center}
	\caption{\label{tab:tab4}Ablation study on the choices of the loss functions for our SA-GAN on ISTD testing set.}
\end{table}
\begin{table}
	\begin{center}
        \resizebox{.45\textwidth}{!}{%
	\begin{tabular}{|c|c|c|c|c|}
			\hline
			Methods & RMSE\(  \downarrow \) & PSNR\(  \uparrow \) & SSIM\(  \uparrow \) & LPIPS\(  \downarrow \) \\
			\hline
			Ours w/o MSP & 6.40 & 27.16 & 0.917 & 0.046 \\
                Ours & \textbf{6.29} & \textbf{27.94} & 0.930 &  \textbf{0.038} \\
                \hline
                Msak-ShadowGAN & 6.96 &25.72 & 0.925 & 0.064 \\
			Mask-ShadowGAN w/ MSP & 6.73 & 26.19 & 0.933 & 0.060 \\
			\hline
                G2R-ShadowNet & 7.33 & 25.72 & 0.941 & 0.047 \\
			G2R-ShadowNet w/ MSP & 6.83 & 25.87 & \textbf{0.952} & 0.053 \\
                \hline
		\end{tabular} % 表格内容
        }     
	\end{center}
	\caption{\label{tab:tab5}Ablation studies on the effectiveness of the MSP on the ISTD testing set.}
\end{table}

As show in Table~\ref{tab:tab3}, we observe that performances of our method reduces across all metrics except SSIM when DBRM is omitted. Comparing row 2 to row 4, we find that the S2F branch is crucial, providing substantial performance gains in terms of all the metrics. Then, when MSP is excluded, there is also a certain drop in performance. In Table~\ref{tab:tab4}, we find that $\mathcal{L}_{GAN_s}$ is important and brings performance improvement. Rows 2 and 5 show a significant decline in SG-GAN's performance without $\mathcal{L}_{GAN_r}$ and $\mathcal{L}_{identity}$, especially for the $\mathcal{L}_{identity}$. Then, when $\mathcal{L}_{\textit{clip}}$, $\mathcal{L}_{cycle}$, and $\mathcal{L}_{shadow}$ is respectively removed from the total loss, the performance of SG-GAN shows a slight decrease. As shown in Figure~\ref{fig:ex6} and Figure~\ref{fig:ex5}, the qualitative results are generally consistent with the aforementioned quantitative results in demonstrating the effectiveness of each component. Compared to the model trained with all components, other variants trained with subsets of these components may exhibit noticeable artifacts in the results. For instance, the result of SG-GAN w/o $\mathcal{L}_{\textit{shadow}}$ in Figure~\ref{fig:ex5} and the result of w/o S2F in Figure~\ref{fig:ex6}, artifact in the shadow region is very prominent.

%------------------------------------------------------------------------
\subsection{Effectiveness of general-purpose MSP} 
We propose a general MSP to mitigate the impact of synthetic images on the performance of the shadow removal network. We experimentally validate the effectiveness of MSP on the ISTD dataset. We conduct experiments not only on our own model but also by integrating our proposed MSP into two unsupervised methods, G2R-ShadowNet \cite{liu2021shadow} and Mask-ShadowGAN\cite{2019Mask}, which require synthetic shadows for shadow removal network training. 

Quantitative results are shown in Table~\ref{tab:tab5}. The results clearly demonstrate that the performance of our method and the other two methods has shown varying degrees of improvement after integrating MSP, particularly in the case of G2R-ShadowNet \cite{liu2021shadow}. It is noteworthy that both our method and Mask-ShadowGAN \cite{2019Mask} are trained in RGB space, whereas G2R-ShadowNet \cite{liu2021shadow} is trained in LAB space. This outcome indicates the universal applicability of our MSP across different color spaces.

%-------------------------------------------------------------------------
\section{Conclusion}
In summary, we propose a novel adversarial diffusion framework for self-supervised shadow removal. Our method consists of two stages: a coarse processing stage and a refined restoration stage, with the main networks being SG-GAN and DBRM, respectively. In SG-GAN, shadows are first generated and then removed, constructing paired training sets used for refinement in DBRM. Additionally, we design a general-purpose MSP module to mitigate the impact of synthetic data on network performance. The coarse results are refined by the powerful generative capabilities of DBRM, restoring texture details and edge artifacts in shadow regions. Extensive experiments demonstrate the effectiveness of our proposed method, showing competitive performance on the ISTD and AISTD datasets compared to other state-of-the-art techniques.
%-------------------------------------------------------------------------
% bibtex
\bibliographystyle{eg-alpha-doi} 
\bibliography{egbibsample}       

\newcommand{\etalchar}[1]{$^{#1}$}
\begin{thebibliography}{\uppercase{SDWMG15}}

\bibitem[AHO11]{2011Shadow}
\textsc{Arbel E., Hel-Or H.}:
\newblock Shadow removal using intensity surfaces and texture anchor points.
\newblock \emph{IEEE Computer Society}, 6 (2011).

\bibitem[And82]{anderson1982reverse}
\textsc{Anderson B.~D.}:
\newblock Reverse-time diffusion equation models.
\newblock \emph{Stochastic Processes and their Applications 12}, 3 (1982), 313--326.

\bibitem[CCZS22]{chen2022simple}
\textsc{Chen L., Chu X., Zhang X., Sun J.}:
\newblock Simple baselines for image restoration.
\newblock In \emph{European conference on computer vision} (2022), Springer, pp.~17--33.

\bibitem[CLH{\etalchar{*}}24]{chen2024symbolic}
\textsc{Chen X., Liang C., Huang D., Real E., Wang K., Pham H., Dong X., Luong T., Hsieh C.-J., Lu Y., et~al.}:
\newblock Symbolic discovery of optimization algorithms.
\newblock \emph{Advances in Neural Information Processing Systems 36} (2024).

\bibitem[CPS20]{cun2020towards}
\textsc{Cun X., Pun C.-M., Shi C.}:
\newblock Towards ghost-free shadow removal via dual hierarchical aggregation network and shadow matting gan.
\newblock In \emph{Proceedings of the AAAI Conference on Artificial Intelligence} (2020), vol.~34, pp.~10680--10687.

\bibitem[DBMH{\etalchar{*}}22]{de2022riemannian}
\textsc{De~Bortoli V., Mathieu E., Hutchinson M., Thornton J., Teh Y.~W., Doucet A.}:
\newblock Riemannian score-based generative modelling.
\newblock \emph{Advances in Neural Information Processing Systems 35} (2022), 2406--2422.

\bibitem[FHLD06]{2006On}
\textsc{Finlayson G.~D., Hordley S.~D., Lu C., Drew M.~S.}:
\newblock On the removal of shadows from images.
\newblock \emph{IEEE Transactions on Pattern Analysis and Machine Intelligence 28}, 1 (2006), 59--68.

\bibitem[FSZY24]{fu2024noise}
\textsc{Fu Z., Song K., Zhou L., Yang Y.}:
\newblock Noise-aware image captioning with progressively exploring mismatched words.
\newblock In \emph{Proceedings of the AAAI Conference on Artificial Intelligence} (2024), vol.~38, pp.~12091--12099.

\bibitem[FY19]{fang2019moving}
\textsc{Fang L., Yu F.}:
\newblock Moving object detection algorithm based on removed ghost and shadow visual background extractor.
\newblock \emph{Laser \& Optoelectronics Progress 56}, 13 (2019), 131002.

\bibitem[FZG{\etalchar{*}}21]{fu2021auto}
\textsc{Fu L., Zhou C., Guo Q., Juefei-Xu F., Yu H., Feng W., Liu Y., Wang S.}:
\newblock Auto-exposure fusion for single-image shadow removal.
\newblock In \emph{Proceedings of the IEEE/CVF conference on computer vision and pattern recognition} (2021), pp.~10571--10580.

\bibitem[GDH13]{2013Paired}
\textsc{Guo R., Dai Q., Hoiem D.}:
\newblock Paired regions for shadow detection and removal.
\newblock \emph{IEEE Transactions on Pattern Analysis and Machine Intelligence 35}, 12 (2013), 2956--2967.

\bibitem[GWY{\etalchar{*}}23a]{guo2023shadowdiffusion}
\textsc{Guo L., Wang C., Yang W., Huang S., Wang Y., Pfister H., Wen B.}:
\newblock Shadowdiffusion: When degradation prior meets diffusion model for shadow removal.
\newblock In \emph{2023 IEEE/CVF Conference on Computer Vision and Pattern Recognition (CVPR)} (2023), IEEE, pp.~14049--14058.

\bibitem[GWY{\etalchar{*}}23b]{guo2023boundary}
\textsc{Guo L., Wang C., Yang W., Wang Y., Wen B.}:
\newblock Boundary-aware divide and conquer: A diffusion-based solution for unsupervised shadow removal.
\newblock In \emph{Proceedings of the IEEE/CVF International Conference on Computer Vision} (2023), pp.~13045--13054.

\bibitem[HJA20]{ho2020denoising}
\textsc{Ho J., Jain A., Abbeel P.}:
\newblock Denoising diffusion probabilistic models.
\newblock \emph{Advances in neural information processing systems 33} (2020), 6840--6851.

\bibitem[HJFH19]{2019Mask}
\textsc{Hu X., Jiang Y., Fu C.~W., Heng P.~A.}:
\newblock Mask-shadowgan: Learning to remove shadows from unpaired data.
\newblock In \emph{2019 IEEE/CVF International Conference on Computer Vision (ICCV)} (2019).

\bibitem[IZZE17]{isola2017image}
\textsc{Isola P., Zhu J.-Y., Zhou T., Efros A.~A.}:
\newblock Image-to-image translation with conditional adversarial networks.
\newblock In \emph{Proceedings of the IEEE conference on computer vision and pattern recognition} (2017), pp.~1125--1134.

\bibitem[JST21]{jin2021dc}
\textsc{Jin Y., Sharma A., Tan R.~T.}:
\newblock Dc-shadownet: Single-image hard and soft shadow removal using unsupervised domain-classifier guided network.
\newblock In \emph{Proceedings of the IEEE/CVF International Conference on Computer Vision} (2021), pp.~5027--5036.

\bibitem[KMP{\etalchar{*}}24]{kubiak2024s3r}
\textsc{Kubiak N., Mustafa A., Phillipson G., Jolly S., Hadfield S.}:
\newblock S3r-net: A single-stage approach to self-supervised shadow removal.
\newblock \emph{arXiv preprint arXiv:2404.12103} (2024).

\bibitem[KOY22]{kim2022diffusion}
\textsc{Kim B., Oh Y., Ye J.~C.}:
\newblock Diffusion adversarial representation learning for self-supervised vessel segmentation.
\newblock \emph{arXiv preprint arXiv:2209.14566} (2022).

\bibitem[LDR{\etalchar{*}}22]{lugmayr2022repaint}
\textsc{Lugmayr A., Danelljan M., Romero A., Yu F., Timofte R., Van~Gool L.}:
\newblock Repaint: Inpainting using denoising diffusion probabilistic models.
\newblock In \emph{Proceedings of the IEEE/CVF conference on computer vision and pattern recognition} (2022), pp.~11461--11471.

\bibitem[LKX{\etalchar{*}}24]{liu2024recasting}
\textsc{Liu Y., Ke Z., Xu K., Liu F., Wang Z., Lau R.~W.}:
\newblock Recasting regional lighting for shadow removal.
\newblock \emph{arXiv e-prints} (2024), arXiv--2402.

\bibitem[LLK23]{lu2023tf}
\textsc{Lu S., Liu Y., Kong A. W.-K.}:
\newblock Tf-icon: Diffusion-based training-free cross-domain image composition.
\newblock In \emph{Proceedings of the IEEE/CVF International Conference on Computer Vision} (2023), pp.~2294--2305.

\bibitem[LRJ{\etalchar{*}}23]{li2023diffusion}
\textsc{Li X., Ren Y., Jin X., Lan C., Wang X., Zeng W., Wang X., Chen Z.}:
\newblock Diffusion models for image restoration and enhancement--a comprehensive survey.
\newblock \emph{arXiv preprint arXiv:2308.09388} (2023).

\bibitem[LS18]{li2018learning}
\textsc{Li Z., Snavely N.}:
\newblock Learning intrinsic image decomposition from watching the world.
\newblock In \emph{Proceedings of the IEEE conference on computer vision and pattern recognition} (2018), pp.~9039--9048.

\bibitem[LS19]{le2019shadow}
\textsc{Le H., Samaras D.}:
\newblock Shadow removal via shadow image decomposition.
\newblock In \emph{Proceedings of the IEEE/CVF International Conference on Computer Vision} (2019), pp.~8578--8587.

\bibitem[LS21]{le2021physics}
\textsc{Le H., Samaras D.}:
\newblock Physics-based shadow image decomposition for shadow removal.
\newblock \emph{IEEE Transactions on Pattern Analysis and Machine Intelligence 44}, 12 (2021), 9088--9101.

\bibitem[LWL{\etalchar{*}}24]{lu2024mace}
\textsc{Lu S., Wang Z., Li L., Liu Y., Kong A. W.-K.}:
\newblock Mace: Mass concept erasure in diffusion models.
\newblock \emph{arXiv preprint arXiv:2403.06135} (2024).

\bibitem[LYM{\etalchar{*}}21]{liu2021Lg}
\textsc{Liu Z., Yin H., Mi Y., Pu M., Wang S.}:
\newblock Shadow removal by a lightness-guided network with training on unpaired data.
\newblock \emph{IEEE Transactions on Image Processing 30} (2021), 1853--1865.

\bibitem[LYW{\etalchar{*}}21]{liu2021shadow}
\textsc{Liu Z., Yin H., Wu X., Wu Z., Mi Y., Wang S.}:
\newblock From shadow generation to shadow removal.
\newblock In \emph{Proceedings of the IEEE/CVF conference on computer vision and pattern recognition} (2021), pp.~4927--4936.

\bibitem[LZB{\etalchar{*}}22]{lu2022dpm}
\textsc{Lu C., Zhou Y., Bao F., Chen J., Li C., Zhu J.}:
\newblock Dpm-solver: A fast ode solver for diffusion probabilistic model sampling in around 10 steps.
\newblock \emph{Advances in Neural Information Processing Systems 35} (2022), 5775--5787.

\bibitem[MFL{\etalchar{*}}24]{mei2024latent}
\textsc{Mei K., Figueroa L., Lin Z., Ding Z., Cohen S., Patel V.~M.}:
\newblock Latent feature-guided diffusion models for shadow removal.
\newblock In \emph{Proceedings of the IEEE/CVF Winter Conference on Applications of Computer Vision} (2024), pp.~4313--4322.

\bibitem[PGM{\etalchar{*}}19]{paszke2019pytorch}
\textsc{Paszke A., Gross S., Massa F., Lerer A., Bradbury J., Chanan G., Killeen T., Lin Z., Gimelshein N., Antiga L., et~al.}:
\newblock Pytorch: An imperative style, high-performance deep learning library.
\newblock \emph{Advances in neural information processing systems 32} (2019).

\bibitem[QTH{\etalchar{*}}17a]{2017DeshadowNet}
\textsc{Qu L., Tian J., He S., Tang Y., Lau R. W.~H.}:
\newblock Deshadownet: A multi-context embedding deep network for shadow removal.
\newblock In \emph{2017 IEEE Conference on Computer Vision and Pattern Recognition (CVPR)} (2017).

\bibitem[QTH{\etalchar{*}}17b]{luo2023image}
\textsc{Qu L., Tian J., He S., Tang Y., Lau R. W.~H.}:
\newblock Deshadownet: A multi-context embedding deep network for shadow removal.
\newblock In \emph{2017 IEEE Conference on Computer Vision and Pattern Recognition (CVPR)} (2017).

\bibitem[RBL{\etalchar{*}}22]{rombach2022high}
\textsc{Rombach R., Blattmann A., Lorenz D., Esser P., Ommer B.}:
\newblock High-resolution image synthesis with latent diffusion models.
\newblock In \emph{Proceedings of the IEEE/CVF conference on computer vision and pattern recognition} (2022), pp.~10684--10695.

\bibitem[RKH{\etalchar{*}}21]{radford2021learning}
\textsc{Radford A., Kim J.~W., Hallacy C., Ramesh A., Goh G., Agarwal S., Sastry G., Askell A., Mishkin P., Clark J., et~al.}:
\newblock Learning transferable visual models from natural language supervision.
\newblock In \emph{International conference on machine learning} (2021), PMLR, pp.~8748--8763.

\bibitem[SCC{\etalchar{*}}22]{saharia2022palette}
\textsc{Saharia C., Chan W., Chang H., Lee C., Ho J., Salimans T., Fleet D., Norouzi M.}:
\newblock Palette: Image-to-image diffusion models.
\newblock In \emph{ACM SIGGRAPH 2022 conference proceedings} (2022), pp.~1--10.

\bibitem[SDWMG15]{sohl2015deep}
\textsc{Sohl-Dickstein J., Weiss E., Maheswaranathan N., Ganguli S.}:
\newblock Deep unsupervised learning using nonequilibrium thermodynamics.
\newblock In \emph{International conference on machine learning} (2015), PMLR, pp.~2256--2265.

\bibitem[SHC{\etalchar{*}}22]{saharia2022image}
\textsc{Saharia C., Ho J., Chan W., Salimans T., Fleet D.~J., Norouzi M.}:
\newblock Image super-resolution via iterative refinement.
\newblock \emph{IEEE transactions on pattern analysis and machine intelligence 45}, 4 (2022), 4713--4726.

\bibitem[SSDK{\etalchar{*}}20]{song2020score}
\textsc{Song Y., Sohl-Dickstein J., Kingma D.~P., Kumar A., Ermon S., Poole B.}:
\newblock Score-based generative modeling through stochastic differential equations.
\newblock \emph{arXiv preprint arXiv:2011.13456} (2020).

\bibitem[WLY18]{2018Stacked}
\textsc{Wang J., Li X., Yang J.}:
\newblock Stacked conditional generative adversarial networks for jointly learning shadow detection and shadow removal.
\newblock In \emph{2018 IEEE/CVF Conference on Computer Vision and Pattern Recognition} (2018).

\bibitem[XHL23]{xu2023learning}
\textsc{Xu K., Hancke G.~P., Lau R.~W.}:
\newblock Learning image harmonization in the linear color space.
\newblock In \emph{Proceedings of the IEEE/CVF International Conference on Computer Vision} (2023), pp.~12570--12579.

\bibitem[XKV21]{xiao2021tackling}
\textsc{Xiao Z., Kreis K., Vahdat A.}:
\newblock Tackling the generative learning trilemma with denoising diffusion gans.
\newblock \emph{arXiv preprint arXiv:2112.07804} (2021).

\bibitem[XSXM13]{2013Fast}
\textsc{Xiao C., She R., Xiao D., Ma K.~L.}:
\newblock Fast shadow removal using adaptive multi‐scale illumination transfer.
\newblock \emph{Computer Graphics Forum 32}, 8 (2013), 207--218.

\bibitem[YWZ{\etalchar{*}}22]{yang2022exploiting}
\textsc{Yang Y., Wei H., Zhu H., Yu D., Xiong H., Yang J.}:
\newblock Exploiting cross-modal prediction and relation consistency for semisupervised image captioning.
\newblock \emph{IEEE Transactions on Cybernetics 54}, 2 (2022), 890--902.

\bibitem[YYB{\etalchar{*}}21]{yang2021corporate}
\textsc{Yang Y., Yang J.-Q., Bao R., Zhan D.-C., Zhu H., Gao X.-R., Xiong H., Yang J.}:
\newblock Corporate relative valuation using heterogeneous multi-modal graph neural network.
\newblock \emph{IEEE Transactions on Knowledge and Data Engineering 35}, 1 (2021), 211--224.

\bibitem[YZG{\etalchar{*}}22]{yang2022domfn}
\textsc{Yang Y., Zhang J., Gao F., Gao X., Zhu H.}:
\newblock Domfn: A divergence-orientated multi-modal fusion network for resume assessment.
\newblock In \emph{Proceedings of the 30th ACM International Conference on Multimedia} (2022), pp.~1612--1620.

\bibitem[ZCDH24]{Zhaowfdiff}
\textsc{Zhao C., Cai W., Dong C., Hu C.}:
\newblock Wavelet-based fourier information interaction with frequency diffusion adjustment for underwater image restoration.
\newblock In \emph{Proceedings of the IEEE/CVF Conference on Computer Vision and Pattern Recognition (CVPR)} (June 2024).

\bibitem[ZCHY24]{zhao2024cycle}
\textsc{Zhao C., Cai W., Hu C., Yuan Z.}:
\newblock Cycle contrastive adversarial learning with structural consistency for unsupervised high-quality image deraining transformer.
\newblock \emph{Neural Networks} (2024), 106428.

\bibitem[ZLM{\etalchar{*}}24]{zhou2024migc}
\textsc{Zhou D., Li Y., Ma F., Yang Z., Yang Y.}:
\newblock Migc: Multi-instance generation controller for text-to-image synthesis.
\newblock \emph{arXiv preprint arXiv:2402.05408} (2024).

\bibitem[ZLY{\etalchar{*}}20]{2020CLA}
\textsc{Zhang L., Long C., Yan Q., Zhang X., Xiao C.}:
\newblock Cla‐gan: A context and lightness aware generative adversarial network for shadow removal.
\newblock \emph{Computer Graphics Forum} (2020).

\bibitem[ZXF{\etalchar{*}}22]{zhu2022efficient}
\textsc{Zhu Y., Xiao Z., Fang Y., Fu X., Xiong Z., Zha Z.-J.}:
\newblock Efficient model-driven network for shadow removal.
\newblock In \emph{Proceedings of the AAAI conference on artificial intelligence} (2022), vol.~36, pp.~3635--3643.

\bibitem[ZYY23]{zhou2023pyramid}
\textsc{Zhou D., Yang Z., Yang Y.}:
\newblock Pyramid diffusion models for low-light image enhancement.
\newblock \emph{arXiv preprint arXiv:2305.10028} (2023).

\end{thebibliography}

% biblatex with biber
%\printbibliography                

%-------------------------------------------------------------------------
%Color tables are no longer required for purely electronic publications.

\end{document}